# A Novel Granular-Based Bi-Clustering Method of Deep Mining the Co-Expressed Genes

Kaijie Xu[1,2], Witold Pedrycz[2,1,5], Zhiwu Li[1,3], Yinghui Quan[1], & Weike Nie[4]

**Traditional clustering methods are limited when dealing with huge and heterogeneous groups of gene expression data, which motivates the development of bi-clustering methods. Bi-clustering methods are used to mine bi-clusters whose subsets of samples (genes) are co-regulated under their test conditions. Studies show that mining bi-clusters of consistent trends and trends with similar degrees of fluctuations from the gene expression data is essential in bioinformatics research. Unfortunately, traditional bi-clustering methods are not fully effective in discovering such bi-clusters. Therefore, we propose a novel bi-clustering method by involving here the theory of Granular Computing. In the proposed scheme, the gene data matrix, considered as a group of time series, is transformed into a series of ordered information granules. With the information granules we build a characteristic matrix of the gene data to capture the fluctuation trend of the expression value between consecutive conditions to mine the ideal bi-clusters. The experimental results are in agreement with the theoretical analysis, and show the excellent performance of the proposed method.**

Gene expression data analysis[1] through DeoxyriboNucleic Acid (DNA) microarray technology helps us solve a variety of problems, such as those encountered in medical diagnosis, molecular biology, and gene expression profiling[2]. So far, numerous technologies have been proposed for discovering such gene regulations. Among them clustering is popular and useful. After the first gene expression datasets became available, clustering remains widely used nowadays[3]. Traditional (global) clustering methods only analyze genes under all experimental conditions or only analyze conditions of all the genes. In practice, in numerous cellular processes, many genes are regularly co-expressed (co-regulated) under some special conditions[4] but behave differently under different conditions. Consequently, mining local co-expressed valuable patterns becomes a vital objective in discovering genetic pathways that are not very clear when clustered globally[5]. Especially with the increasing number of conditions most of the objects in a dataset tend to be nearly equidistant from each other, completely masking the clustering structure. Then traditional (distance-based measures) clustering methods are unable to deal with this aspect, which motivates the development of the bi-clustering methods. Bi-clustering is not only able to reveal the global structure in data but it can discover the local information (in other words, it can form clusters in the feature space and the data space simultaneously).

Bi-clustering, first introduced by Hartigan[6], has been further developed since Cheng and Church proposed a bi-clustering method (CC method) based on variance and applied it to gene expression data[7]. Their work remains one of the most important contributions to the bi-clustering field. Bi-clustering methods can uncover co-expressed valuable patterns of gene from plenty of gene expression data, which are more helpful to define genes functioning together than traditional clustering approaches. Popular bi-clustering methods, such as CC[7], Flexible Overlapped Clusters (FLOC)[8], Plaid[9], order-preserving sub-matrix (OPSM)[10], Iterative Signature Algorithm (ISA)[11], Spectral bi-clustering method[12], conserved gene expression MOTIFs (xMOTIFs)[13], and BiMax[14] have drawn much attention in the literature. Several other methods of bi-clustering using various techniques have also been reported, such as those based on evolutionary methods[15], hierarchical bi-clustering[16], and Bayesian bi-clustering[17]. The CC method uses mean squared residue of a bi-cluster as a similarity measure to greedily extract bi-clusters that satisfy a homogeneity constraint. Based on the CC and with the aim of improving the generic CC method, Yang et al. proposed another algorithm called Flexible Overlapped Clusters method[8], where an additional function is introduced to deal with the missing data and to discover the overlapping bi-clusters[18]. Both the CC and the Flexible Overlapped Clusters methods optimizing the mean squared residue have been considered to be the most effective tools for processing gene expression data.

Mean squared residue is the most commonly used index in bi-clustering, however, it cannot always capture the trend consistency of bi-clusters accurately. Patterns with larger mean squared residue scores may present more consistent trends than those with smaller ones[19]. In contrast, some patterns with smaller mean squared residue scores do not exhibit visible consistent trends. The order-preserving sub-matrix method[10], focusing on the relative order of the expression levels of each gene rather than the exact values, is proposed to find the hidden patterns in gene expression data. However, the method only concerns the size of the data, but ignores the actual value. Meanwhile, the order-preserving sub-matrix method requires more computing resources. Several methods based on matrix transformation were proposed to discover the bi-clusters with consistent trends, such as Deterministic Bi-clustering with

[1]Xidian University, Xi'an, China. [2]Department of Electrical and Computer Engineering, University of Alberta, Edmonton, Canada. [3]Institute of Systems Engineering, Macau University of Science and Technology, Taipa, Macau, China. [4]School of Information Science and Technology, Northwest University, Xi'an, China. [5]Faculty of Engineering, King Abdulaziz University, Jeddah, Saudi Arabia.

Frequent pattern mining[20], Quick Hierarchical Bi-clustering method[19], and SKeleton Bi-clustering[21]. These methods can produce some more meaningful mining results. However, they still rely only on numerical data, but ignore their semantic aspects (say, linkages among data) and some hidden information is not discovered.

Granular Computing[22] is a computing paradigm and emerging conceptual framework of information processing and plays an important role in many areas. Especially, in the field of Computational Intelligence, Granular Computing can model human reasoning to help deal with complex problems.

This study intends to introduce the theory of Granular Computing to the field of the DNA microarray technology. In the paper, we focus on building information granules to interpret the gene expression data, which will make the data easier to understand. More specifically, in this study, the gene data matrix is considered as a time series group that is transformed into a series of ordered information granules. Information granularity inherently arises when it comes to the interpretation and perception of time series (gene expression data). Using a vocabulary of information granules to describe the time series makes the interpretation of data easier to comprehend.

To build information granules, we use the time series theory in conjunction with the Fuzzy C-Means (FCM) clustering[23], and the principle of justifiable information granularity[24]. Then we label the conditions of each gene based on the information granules to construct a characteristic matrix of the gene data matrix, which can capture and supervise the changing trend of the gene expression value between consecutive conditions. A collection of initial bi-clusters is generated by traversing the trend matrix and utilizing the D-miner method[25]. In the final phase, the initial bi-clusters are refined with the supervision of mean squared residue and mean fluctuation degree[19]. All these features contribute to the originality of the proposed approach. To quantify the performance of the method, a detailed analysis and a comprehensive suite of experiments are provided. The experimental studies demonstrate that the proposed approach achieves better performance compared with that of the two well-known methods used for gene expression. To the best of our knowledge, the idea of the proposed approach has not been exposed in the previous studies.

## BUILDING THE ORDERED INFORMATION GRANULES

In this section, we build a group of ordered information granules to explain the microarray expression data. In this process, each gene with all conditions is considered as a time series at different time points. Then the series are granulated into several ordered information granules. The conditions of each gene will be labeled according to the information granules.

Let $g_i = [g_1, g_2, ..., g_M]$ be a gene with $M$ conditions, which is considered to be a multivariable time series $g(t)$ ($t = 1, 2, ..., M$).

Before proceeding with the granulation process, we sort the numerical data in an ascending order and treat the data with the same values as a single datum. This phase forms an important step in the proposed method; Fig. 1 shows the details of data processing. Assume that the sorted data are as follows:

$$g_{1,0} = g_{1,1} = \cdots < g_{2,0} = g_{2,1} = \cdots < \cdots < g_{t,0} = g_{t,1} = \cdots g_{t,m} \cdots < \cdots \quad (1)$$

where $g_{t,0}, g_{t,1}, ...,$ and $g_{t,m}$ denote the data with the same value. Assume that the sorted data sequence $g'(t)$ contains $t$ ($t \leqq M$) different numerical data, and in the proposed method the building of information granules for $g(t)$ needs the assistance of the $g'(t)$.

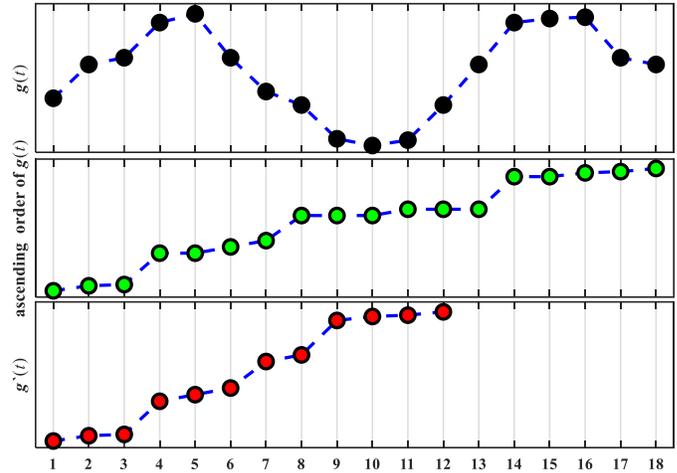

Fig. 1. Example of data processing; see the description in the text

Before building the information granules, some special cases should be considered. In the gene expression matrix such data occasionally occur:

$$\begin{aligned} \text{case 1:} \quad & g_1 = g_2 = \cdots = g_M \text{ namely, card}\{g'(t)\} = 1 \\ \text{case 2:} \quad & \text{card}\{g'(t)\} = 2 \end{aligned} \quad (2)$$

When encountering such cases as shown above, instead of implementing the following granulation methods, the information granules are presented as interval information granules to cover the data in the following way:

$$\begin{aligned} \text{case 1:} \quad & \boldsymbol{\Omega}_1 = [g_1, g_1) \\ \text{case 2:} \quad & \boldsymbol{\Omega}_1 = [\min[g'(t)], (\max[g'(t)] + \min[g'(t)])/2) \\ & \boldsymbol{\Omega}_2 = [(\max[g'(t)] + \min[g'(t)])/2, \max[g'(t)]] \end{aligned} \quad (3)$$

Otherwise, the methods of building information granules will be implemented.

*Building Information Granules Based on Fuzzy C-Means clustering*

With the FCM clustering, the structure of available numerical data is represented by a partition (membership) matrix and prototypes (cluster centers). Considering the time series vector $g(t)$ to be granulated into $C$ information granules, a time series $g(t)$ is ultimately expressed as

$$\left[ QU^\kappa \right] g = V \quad (4)$$

where $U = [\cdots, \mu_{tj}, \cdots] \in R^{C \times M}$ is the partition matrix, $\mu_{tj}$ is the degree of membership of sample point $g_t$ belonging to the $j$th cluster, $\kappa$ ($\kappa > 1$) is a fuzziness coefficient[23], $Q$ is such a diagonal matrix

$$Q = \text{diag}\{\cdots, 1 / \sum_{t=1}^{M} \mu_{tj}^\kappa, \cdots\} \quad (5)$$

and $V \in R^{K \times 1}$ is a collection of the prototypes with $K$ being the number of clusters. Then, we construct an augmented matrix in

the form
$$Z = [U \ V] \quad (6)$$
To obtain the ordered information granules, we sort all other columns of the matrix according to the ascending order of the last column, and label them also in an ascending order. The sorted augmented matrix is
$$Z' = [U' \ V'] \quad (7)$$
Thus, the ordered information granules can be determined by $U'$ and $V'$. In addition, we set the number of the information granules as
$$C = \text{ceil}\left[0.5 \text{card}\{g'(t)\}\right] \quad (8)$$
where $\text{ceil}[x]$ denotes the least integer value greater than or equal to $x$.

*Building Information Granules Based on The Principle of Justifiable Information Granularity*

Building information granules based on the principle of justifiable information granularity is carried out on the time series $g'(t)$. Here, the information granules describe the characteristics of the distributions of the gene expression values that are used to build a trend matrix of the gene expression data in the next phase of the proposed method to capture the fluctuation trend of the gene expression values. The specific process is as follows:

Partition the time series $g'(t)$ into $P$ sub-series temporal windows $T_1, T_2, \ldots, T_P$, and denote the corresponding sub-time series are $S_1, S_2, \ldots, S_P$. For each sub-time series by $S_p$, we build an optimized information granule $\Omega_p^\tau = [\alpha_p, \beta_p]$ under a given information granularity level $\tau$, where $\tau$ is a positive parameter delivering some flexibility[24], which controls the justifiable granularity (coverage) when optimizing the interval information granule $\Omega$. In the process, two intuitively compelling requirement (criteria) are considered: one is justifiable granularity (coverage) and the other is well-articulated semantics. Since the two criteria are in conflict (the increase of coverage reduces a level of flexibility and vice versa), an aggregate performance index in the form of the product of coverage and specificity is sought. By maximizing it, one produces the lower and upper bound of the interval information granule[24], namely

$$\max: V(\beta) = f_1(\text{card}\{g'(t) \in S_p \mid \text{med}(S_p) < g'(t) \leq \beta\}) \\ * f_2(|\text{med}(S_p) - \beta|) \quad (9)$$

$$\max: V(\alpha) = f_1(\text{card}\{g'(t) \in S_p \mid \alpha \leq g'(t) < \text{med}(S_p)\}) \\ * f_2(|\text{med}(S_p) - \alpha|) \quad (10)$$

where $\text{med}(S_p)$ is the median of $S_p$. $f_1$ and $f_2$ are increasing and decreasing functions, respectively. The most frequently used forms of $f_1$ and $f_2$ are
$$f_1(u) = u, \quad f_2(u) = \exp(-\tau u) \quad (11)$$
By maximizing the performance index, a group of optimal lower and upper bounds of an interval can be obtained to generate a number of information granules $\Omega_p \ (p = 1, 2, \cdots, P)$.

Usually, we wish the information granules on the time series are the most "informative" (compact) so that they carry a clearly articulated semantics[26]. In other words, we need to find the optimal partition windows $T_1, T_2, \ldots, T_P$ on the time series $g'(t)$. An overall measure with this regard is to construct and optimize a multivariable objective function to find the optimal segmentation points. For each sub-time series $S_p$, we compute the volume of the associated information granule $\text{Vol}(\Omega_p)$ by

$$\text{Vol}(\Omega_p) = T_p \int_{g'_{\min}}^{g'_{\max}} \Omega_p(g') \quad (12)$$

Note that the bounds ($g'_{\min}$ and $g'_{\max}$) of the integration are the minimal and maximal values of the temporary temporal window $T_p$. By using the methods of global optimization such as Particle Swarm Optimization[27] we minimize the following multivariable objective function to make the sum of the volume of the information granules reach the minimum:

$$\min T_1, T_2, \cdots, T_P \sum_{p=1}^{P} \text{Vol}(\Omega_p) \quad (13)$$

the optimized partition windows $T_1, T_2, \ldots, T_P$ are obtained. The time series $g'(t)$ is transformed into $P$ interval information granules to present more efficient interpretation of the data. Assume that the $P$-1 segmentation points are $g'_1, g'_2, \cdots, g'_{P-1}$ arranged in an ascending order. Then we build $P$ intervals as

$$\Phi_1 = [\min(g'(t)), g'_1], \Phi_2 = [g'_1, g'_2], \Phi_2 = [g'_{P-2}, g'_{P-1}], \cdots, \\ \Phi_P = [g'_{P-1}, \max(g'(t))] \quad (14)$$

So far, the time series $g'(t)$ has been transformed into interval information granules, labeled in an ascending order. Then we label the time series $g(t)$ according to the labels of the information granules.

## GRANULAR-BASED BI-CLUSTERING

Based on the ordered information granules, in this section, we present the granular-based bi-clustering model.

*Construction of a Trend Matrix and a Slope Angle Matrix of the Gene Data Matrix*

Consider a gene expression data matrix $A \in R^{N \times M}$ whose information granular labeling matrix is expressed as
$$\Psi_A = [\cdots, \psi_{ij}, \cdots] \in R^{N \times M} \quad (15)$$
Based on $\Psi_A$, we form a trend matrix of the gene expression matrix as follows:

$$\Gamma_A = \begin{bmatrix} [0]_{11}^{\{1 \times 1\}} & [\cdots]_{12}^{\{1 \times 1\}} & \cdots & [\cdots]_{1m}^{\{1 \times (m-1)\}} & \cdots & [\cdots]_{1M}^{\{1 \times (M-1)\}} \\ \vdots & \vdots & \ddots & \vdots & \ddots & \vdots \\ [0]_{n1}^{\{1 \times 1\}} & [\cdots]_{n2}^{\{1 \times 1\}} & \cdots & [\cdots]_{nm}^{\{1 \times (m-1)\}} & \cdots & [\cdots]_{nM}^{\{1 \times (M-1)\}} \\ \vdots & \vdots & \ddots & \vdots & \ddots & \vdots \\ [0]_{N1}^{\{1 \times 1\}} & [\cdots]_{N2}^{\{1 \times 1\}} & \cdots & [\cdots]_{Nm}^{\{1 \times (m-1)\}} & \cdots & [\cdots]_{NM}^{\{1 \times (M-1)\}} \end{bmatrix} \quad (16)$$

where $\Gamma_A$ contains $N \times M$ sub-matrices, and the values of elements in each sub-matrix are 0, +1 and -1, which characterize all the rising and falling trends of different conditions of genes. It needs to be pointed out that +1 represents a rising trend, -1 represents a falling trend, and 0 represents both the rising and falling trends. The calculation of the sub-matrix with the *n*th row and the *m*th column in $\Gamma_A$ is shown as

$$[\cdots]_{nm}^{\{1 \times (m-1)\}} = [\gamma_{n1}, \cdots, \gamma_{nk}, \cdots, \gamma_{n(m-1)}]; \quad \gamma_{nk} = \text{sign}(\psi_{nm} - \psi_{nk}) \quad (17)$$

With the characteristic matrix $\Gamma_A$, the following task is to find the bi-clusters.

*Construction of Initial Bi-Clusters*

Suppose that we start mining the bi-clusters from the $c$th condition (column). The next ($m$th) condition is determined by the following form

$$\left\{ m \left| \max_{m=c,\cdots,(m+L)} \left[ \begin{array}{l} \sum_{n=1}^{N}\left(\gamma_{nm}=+1 \big| \gamma_{nm} \in [\cdots]_{nm}^{\{1\times(m-1)\}}\right)+ \\ \sum_{n=1}^{N}\left(\gamma_{nm}=0 \big| \gamma_{nm} \in [\cdots]_{nm}^{\{1\times(m-1)\}}\right), \\ \sum_{n=1}^{N}\left(\gamma_{nm}=-1 \big| \gamma_{nm} \in [\cdots]_{nm}^{\{1\times(m-1)\}}\right)+ \\ \sum_{n=1}^{N}\left(\gamma_{nm}=0 \big| \gamma_{nm} \in [\cdots]_{nm}^{\{1\times(m-1)\}}\right) \end{array} \right] \right. \right\} \quad (18)$$

which means that we are searching for the maximum same trends (MSTs) in the next $L$ conditions rather than all of them each time to make the bi-clusters contain as many co-expressed genes as possible. Here the parameter $L$ is used to control the number of genes and conditions in the initial bi-cluster. We set a unit pulse response function $\varepsilon$ to adaptively control the value of $L$ during the search process in such a way:

$$L = L\varepsilon(M - L_0 - c) + (M - c)\varepsilon(L_0 + c - M) \quad (19)$$

where $L_0$ is the initial value of $L$, and the larger the $L_0$ is, the more rows the bi-cluster will contain; the smaller the $L_0$ is, the more columns the bi-cluster will have. Therefore, we obtain different (more) bi-clusters with different values of $L_0$, which is of great significance in gene expression.

When the next condition (the $k$th column) is obtained based on the initial (current, the $c$th) condition, the $k$th column will be set as the current condition to seek the next optimal condition (the ($k$+1)th column). The following search methods are basically the same as the initial one. The difference is that the genes (rows) participating in the following calculation will not be all genes (rows), but the genes (rows) that form the MSTs from the $c$th to the ($c$+$L$)th, that is

$$\left\{ n \left| \max_{m=c,\cdots,(m+L)} \left[ \begin{array}{l} \sum_{n=1}^{N}\left(\gamma_{nm}=+1 \big| \gamma_{nm} \in [\cdots]_{nm}^{\{1\times(m-1)\}}\right)+ \\ \sum_{n=1}^{N}\left(\gamma_{nm}=0 \big| \gamma_{nm} \in [\cdots]_{nm}^{\{1\times(m-1)\}}\right), \\ \sum_{n=1}^{N}\left(\gamma_{nm}=-1 \big| \gamma_{nm} \in [\cdots]_{nm}^{\{1\times(m-1)\}}\right)+ \\ \sum_{n=1}^{N}\left(\gamma_{nm}=0 \big| \gamma_{nm} \in [\cdots]_{nm}^{\{1\times(m-1)\}}\right) \end{array} \right] \right. \right\} \quad (20)$$

Obviously, as the search progresses, the columns (conditions) of the bi-cluster are constantly increasing and rows decreasing at the same time. To make the bi-cluster (sub-matrix) contain more genes (rows), here we set a terminal parameter $min\text{Gene}$, namely, $\text{card}\{n\} \geq min\text{Gene}$. Meanwhile, we also set a parameter $min\text{Cond}$ to make the bi-cluster contain more conditions. Thus, a number of different initial bi-clusters can be obtained without violating the following requirements

$$\text{card}\{n\} \geq min\text{Gene}, \ \text{card}\{m\} \geq min\text{Cond} \quad (21)$$

For the different values of the initial current condition $c$, we can also obtain more different initial bi-clusters, and to construct much more initial bi-clusters we can set the MSTs as the second (or third etc.) MSTs or their combinations in the search process. In addition, by changing the value of $L_0$, the number of different initial bi-clusters can continue to increase dramatically. With the proposed approach each initial bi-cluster exhibits high consistency.

*Refine the Bi-Clusters*

Although a number of bi-clusters have been obtained, they are not always optimal to some extent. In this section, we refine them with the aid of some indexes.

We build another important characteristic matrix of the gene expression data, namely slope angle matrix, defined as

$$\Theta = 180 \arctan\left[ pinv(\Delta)\Xi \right]/\pi \quad (22)$$

where $pinv$ is the pseudo-inverse of a diagonal matrix $\Delta$ expressed as

$$\Delta = diag\left\{ \max(a_{1j}) - \min(a_{1j}), \cdots, \max(a_{ij}) - \min(a_{ij}), \cdots \right\}/(M-1) \quad (23)$$
$$i = 1, 2, \cdots, N; \ j = 1, 2, \cdots, M$$

and $\Xi$ is an adjacent difference matrix that characterizes the differences between any two adjacent conditions in each gene, say,

$$\Xi = [\cdots, a_{ij} - a_{i(j-1)}, \cdots] \quad i = 1, 2, \cdots, N; \ j = 2, \cdots, M \quad (24)$$

With the slope angle matrix, a mean fluctuation degree of a bi-cluster can be defined as

$$MFD(I, J) = \sqrt{\frac{1}{|I||J|} \sum_{i \in I, j \in J}\left( \Theta_{ij} - \frac{1}{|I|}\sqrt{\Theta_{ij}} \right)^2} \quad (25)$$

where $I \subset N$ and $J \subset M$ are subsets of genes and conditions, respectively. Obviously, for a bi-cluster if the genes (rows) have similar changing trends under each condition transition, its mean fluctuation degree score will be relatively smaller. Furthermore, if all genes (rows) in the bi-cluster have the completely similar (or same) changing trends under each condition transition, its mean fluctuation degree score will be zero[19]. Therefore, we can refine the bi-clusters by combining the mean squared residue with the mean fluctuation degree. The refinement process consists of two steps.

**Step 1:** Delete rows and columns. Calculate the mean squared residue and the mean fluctuation degree scores of each initial bi-cluster and the mean squared residue of each row and column. Delete the rows and columns if they satisfy the following two conditions: **i.** the mean squared residue scores are larger than the given threshold; **ii.** after deleting them the mean fluctuation degree scores can be decreased; otherwise, this step will not be implemented.

**Step 2:** Add rows and columns. We add the rows and columns that are not in the bi-cluster and will not increase the mean fluctuation degree scores.

In the processes of deletion and addition of the rows and columns, the mean squared residue and mean fluctuation degree are constantly recalculated and improved (reduced) until a new optimal bi-cluster is obtained.

## EXPERIMENTAL RESULTS

In the following experiments, we compare the performance of the proposed method with CC and the Flexible Overlapped Clusters methods, which are the two well-known bi-clustering methods commonly used for gene expression. The implementation is completed in MATLAB. In the experiments, a commonly used gene expression dataset (Gordon-2002) is used, for a detailed description refer to[28]. The methods are used to find 50 bi-clusters with the mean squared residue threshold of 1,200. In the use of the proposed method, the parameters $min$Gene, $min$Cond and $L_0$ which control the size (number of the co-expressed genes and conditions) of the bi-clusters are set as 15, 10 and 30. For the Fuzzy C-Means clustering, the fuzziness coefficient $\kappa$ is set as 2, which is the most frequently used value in practice. The methods are repeated 10 times; the means and the standard deviations of the experimental results are presented.

The experimental results are plotted in Figs. 2 to 4, where the GBC-I represents the building information granules with Fuzzy C-Means clustering, GBC-II the principle of justifiable information granularity, CC the Cheng and Church's method and the Flexible Overlapped Clusters method. Fig. 3 plots the heat map and 12 bi-clusters of the dataset obtained by all the methods for visualization comparison (Note that these are not average experiments, but the results of one of the experiments), and in the bi-clusters each color curve represents a single gene. We can see that the trend lines of the bi-clusters based on the proposed methods are very neat and orderly. On the contrary, those obtained by the CC and Flexible Overlapped Clusters methods are disorderly, though they contain more genes and conditions. Accordingly, the mean fluctuation degree scores plotted in Fig. 2 show that the proposed methods exhibit smaller mean values of these degrees. Furthermore, both the initial and the refined bi-clusters have good performance of mean fluctuation degree, in other words, the proposed methods are very effective in discovering the quality of initial bi-clusters by grouping together genes that have trends with more similar fluctuation degrees. Although the mean squared residue of the initial bi-clusters are larger than the given threshold, their mean fluctuation degree scores show that they also have the same changing trends. The CC and the Flexible Overlapped Clusters methods have obtained the bi-clusters without violating the requirements (a given mean squared residue threshold), however, they are not well suited for gene expression (not fully effective in discovering co-expressed genes under some special conditions) to some extent.

In addition, the impact of the random numbers in CC method and the random selection of initial seeds in Flexible Overlapped Clusters method make them very unstable, which are reflected in the large standard deviations as shown in Fig. 2. In Fig. 4, we also present a group of co-expressed genes of the dataset obtained by all the methods.

In other words, compared with the CC and Flexible Overlapped Clusters methods, the proposed methods exhibit visible advantages. The improvement is 30% (on average) and varies in-between a minimal improvement of 25% and 35% in the case of the most visible improvement.

For the computing overhead, the proposed method searches the bi-clusters purposefully and obtains the results as fast as possible, which is outperforms the CC and Flexible Overlapped Clusters methods. Building the information granules and calculating the trend matrix of the gene expression data are also time consuming. Fortunately, both procedures are done off-line, and only needs to be implemented once for a given dataset and then stored.

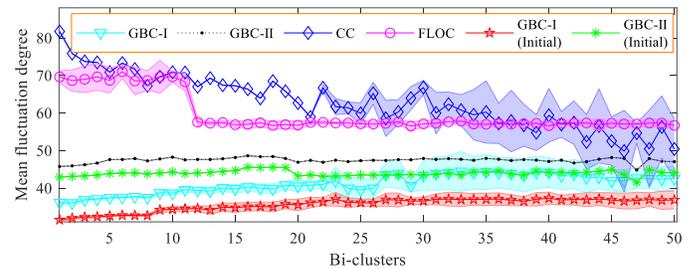

Fig. 2. Comparison of the mean fluctuation degree on the Gordon-2002 dataset

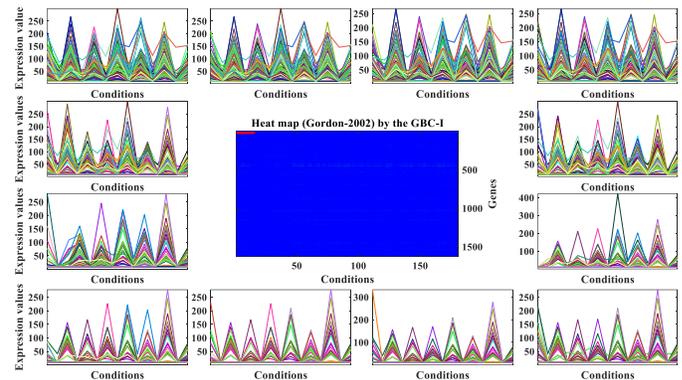

Fig. 2-a. Gene expression profiles of bi-clusters determined by the GBC-Ⅰ.

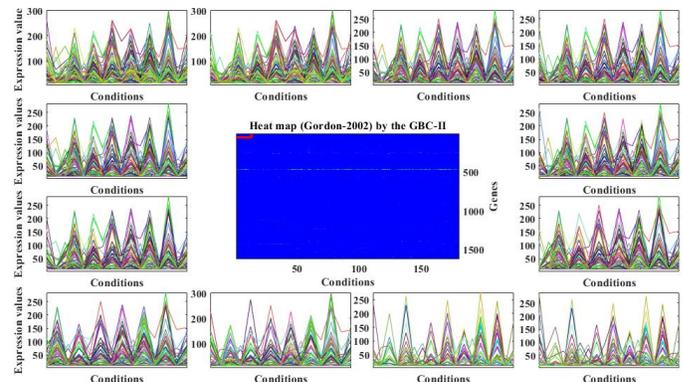

Fig. 3-b. Gene expression profiles of bi-clusters determined by the GBC-Ⅱ.

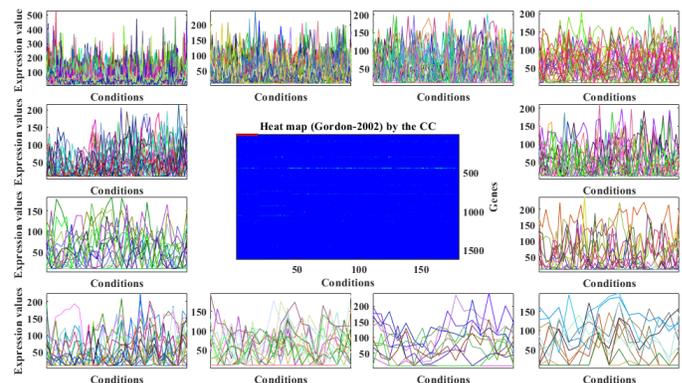

Fig. 3-c Gene expression profiles of bi-clusters determined by the CC.

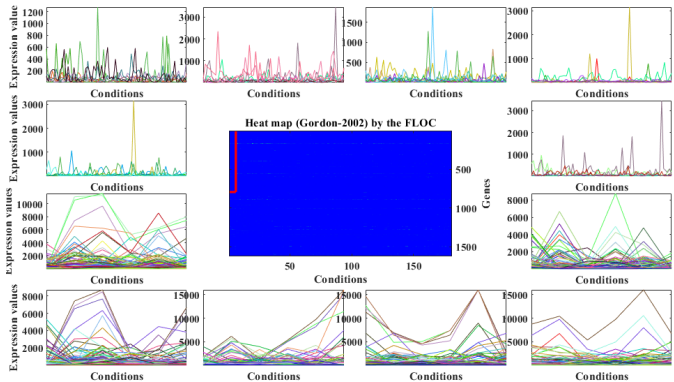

Fig. 3-d Gene expression profiles of bi-clusters determined by the FLOC.

Fig. 4-a Co-expressed genes of a bi-cluster determined by the GBC-Ⅰ.

Fig. 4-b Co-expressed genes of a bi-cluster determined by the GBC-Ⅱ.

Fig. 4-c Co-expressed genes of a bi-cluster determined by the CC.

Fig. 4-d Co-expressed genes of a bi-cluster determined by the FLOC.

## CONCLUSIONS

In this research, we designed a bi-clustering method to discover the co-expressed genes based on the theory of granular computing. During the design process, information granules are introduced to transform the gene numerical data to capture the changing trend of the gene expression value between consecutive conditions. Bi-clusters are obtained through extracting the features of the information granules. We completed theoretical analysis and offered a comprehensive suite of experiments. Both the theoretical and experimental results are presented to verify the validity of the proposed method. Experimental results show that the proposed method outperforms the existing methods in finding the bi-clusters. To the best of our knowledge, this research approach offers an innovative direction to bi-clustering and comes with tangible improvements (say 30% improvement on average has been achieved). While at this phase sound background promising study has been completed, more experimental work could be pursued in the future. A viable and promising alternative would be to engage other formalisms of information granules, say fuzzy sets or rough sets.